\pdfoutput=1
\documentclass[11pt]{article}

\usepackage[]{ACL2023}

\usepackage{times}
\usepackage{latexsym}

\usepackage[T1]{fontenc}
\usepackage[utf8]{inputenc}

\usepackage{microtype}
\usepackage{hyperref}

\usepackage{inconsolata}
\usepackage{graphicx}
\usepackage{multirow}
\usepackage{scrextend}
\usepackage{cjhebrew}

\newcommand{\me}[1]{\textcolor{magenta}{[ME: #1]}}
\newcommand{\he}[1]{\textcolor{violet}{[HE: #1]}}
\newcommand{\ra}[1]{\textcolor{green}{[RA: #1]}}
\newcommand{\rt}[1]{\textcolor{orange}{[RT: #1]}}
\newcommand{\is}[1]{\textcolor{blue}{[IS: #1]}}

\renewcommand{\me}[1]{}
\renewcommand{\he}[1]{}
\renewcommand{\ra}[1]{}
\renewcommand{\rt}[1]{}
\renewcommand{\is}[1]{}

\title{Multilingual Sequence-to-Sequence Models for Hebrew NLP}

\author{Matan Eyal \quad Hila Noga \quad Roee Aharoni \\ 
\bf{Idan Szpektor} \quad \bf{Reut Tsarfaty} \\ \\
        Google Research \\
        \tt\{matane,hilanoga,roeeaharoni,szpektor,reutt\}@google.com}
\begin{document}
\maketitle
\begin{abstract}
Recent work attributes progress in NLP to large language models (LMs) with increased model size and large quantities of pretraining data.
Despite this, current state-of-the-art LMs for Hebrew are both under-parameterized and under-trained compared to LMs in other languages. 
Additionally, previous work on pretrained Hebrew LMs focused on encoder-only models. 
While the encoder-only architecture is beneficial for classification tasks, it does not cater well for sub-word prediction tasks, such as Named Entity Recognition, when considering the morphologically rich nature of Hebrew.
% where words are composed of a variable number of morphemes.
In this paper we argue that sequence-to-sequence generative architectures are more suitable for LLMs in the case of {\em morphologically rich languages} (MRLs) such as Hebrew.
We demonstrate that by casting tasks in the Hebrew NLP pipeline as text-to-text tasks, we can leverage powerful multilingual, pretrained sequence-to-sequence models as mT5, eliminating the need for a specialized, morpheme-based, separately fine-tuned decoder.
Using this approach, our experiments show substantial improvements over previously published results on existing Hebrew NLP benchmarks. 
These results suggest that multilingual sequence-to-sequence models present a  promising building block for NLP for MRLs. %are more effective than previous encoder-only models for Hebrew and other MRLs.
\end{abstract}

\section{Introduction}
\label{introduction}

In recent years, large pretrained language models showed impressive results in a variety of NLP tasks \cite{devlin-etal-2019-bert, raffel2020exploring}, domains \cite{beltagy-etal-2019-scibert, araci2019finbert}, and languages \cite{antoun-etal-2020-arabert, chan-etal-2020-germans}.
These models were trained in a self-supervised fashion on large corpora with language modeling objectives. By doing this, models are taking advantage of information available in the training data without any access to explicit labels \cite{petroni-etal-2019-language}. Recent works \cite{kaplan2020scaling, hoffmann2022training} argue that language models capabilities scale as a power-law with model size, dataset size, and the amount of compute used for training.

\begin{figure}[t]
\centering
\includegraphics[width=\linewidth]{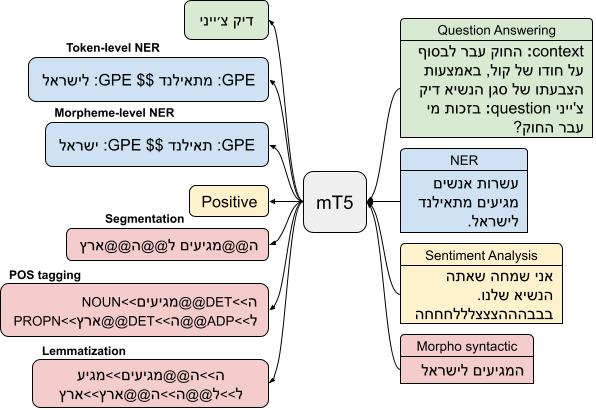}
\caption{Finetuning mT5 on Hebrew tasks in text-to-text settings}
\label{flow}
\end{figure}

After the success such models achieved on English benchmarks, analogous language-specific models were developed to improve benchmark results in a variety of languages such as Arabic and French to name a few \cite{antoun-etal-2020-arabert, martin-etal-2020-camembert}. Hebrew NLP was no different with a number of BERT variations \cite{devlin-etal-2019-bert}  proposed: HeBERT \cite{chriqui2021hebert}, AlephBERT \cite{seker-etal-2022-alephbert}, and recently AlephBERTGimmel (ABG) a variation of AlephBERT with a larger vocabulary \cite{ABG}.

Contrary to the scaling laws proposed in \citet{kaplan2020scaling}, all of the Hebrew variations of BERT are both trained with a relatively small pretraining data, and are under-parameterized.
In terms of training data, HeBERT and AlephBERT (and similarly ABG) were trained on 10.5GB and 16GB of Hebrew pretraining data respectively. mT5 \cite{xue-etal-2021-mt5} in comparison, was trained on mC4, where its public replica\footnote{\href{https://github.com/allenai/allennlp/discussions/5265}{https://github.com/allenai/allennlp/discussions/5265}} is a 27TB collection of natural text in 101 languages drawn from the public Common Crawl, 66GB of which is Hebrew (4.125x more Hebrew training data compared to AlephBERT).
In terms of the model size, ABG, the largest Hebrew LM, has 60x and 945x fewer parameters compared to English T5 XXL \cite{raffel2020exploring} and GPT3 \cite{brown2020language}, respectively, language models that were released two years earlier.
Although English T5 is a much larger model than available Hebrew language models, it can still be fine-tuned on common GPUs. See Appendix \ref{running_t5_on_common_gpus} for more details.
% While being much larger than available Hebrew LMs, T5 can be finetuned on common GPUs, see Appendix \ref{running_t5_on_common_gpus} for further details.

In addition to scale differences, all previous Hebrew LMs use an encoder-only architecture, even though the morphological complexity of Hebrew and other {\em morphologically rich languages} (MRLs)\footnote{MRLs, in contrast configurational languages (e.g. English), express grammatical functions at the word level via phenomena such as inflectional affixes, pronominal clitics, etc.} pose challenges for the effectiveness of this model.
% This model is challenged by  presents difficulties for Morphologically Rich Languages (MRLs)\footnote{MRLs, in contrast configurational languages (e.g. English), express grammatical functions at the word level via phenomena such as inflectional affixes, pronominal clitics, etc.} such as Hebrew, especially when performing tasks like token classification and span prediction.
Consider the task of POS tagging: Assigning POS tags for the phrase \texttt{``babayit halavan"}\footnote{Hebrew transcribed to English. Translated as ``In the White House". The phrase is made of the morphemes be-ha-bayit ha-lavan (in-the-house the-white), but written and pronounced as ``babayit halavan" without explicit  boundaries.} requires to initially segment the phrase to its morphemes and only then assign each morpheme its matching POS tag. 
Since the number of input tokens does not match the number of output tags (2 input words and 5 output tags, one for each morpheme), a one-to-one token-to-tag classification head, as commonly employed in encoder-only models, is not feasible.
The same problem appears in semantic tasks like Question-Answering (QA) and Named Entity Recognition (NER). For example, the named entity for \texttt{``\textbf{ba}bayit halavan"} is \texttt{``\textbf{ha}bayit halavan"}.
This goes beyond what encoder-only models can do by requiring the model to label a string that is not part of the input text.

To overcome this architectural obstacle in encoder-only models, AlephBERT authors used \citet{brusilovsky2022neural} three-step segmentation and tagging approach:
Contextualize the input, pass resulting embeddings to LSTM decoder that generates the segmentation separated by a space symbol, finally pass the white space representation to a classification head.
% First, the input text is is transferred to contextualized representations using the LM encoder. Secondly, each word-piece representation is passed to a char-based LSTM decoder that generates the matching morphemes, separated by a space symbol. Thirdly, the space symbol's representation is then passed to a classifier to pick the most probable tag.
While effective for morpho-syntactic tasks, these additional architecture components do not enable full generartive capabilities and are not pretrained, therefore cannot benefit from some of pretrained LMs advantages.
% Furthermore, while helpful for morpho-syntactic tasks, such architectural change does not enable full generative capabilities relevant for tasks such as abstractive summarization, paraphrasing etc.
% Furthermore, this three-step model cannot enjoy from being trained in an autoregressive nature where at each time step all previous labels (here that is segmentations and labels) are visible. \me{Right? Pretty sure about this}
% \rt{i'd also mention that the errors in this separate segmentation stage may linger and create noise downstream}\me{Will they linger more than in encoder-decoder? Why? I wrote this section but removed becauase of length considerations: Furthermore, this three-step model cannot enjoy from being trained in an autoregressive nature where at each time step all previous labels (here that is segmentations and labels) are visible.}

In contrast, sequence-to-sequence models can simply take the raw text as input and, for POS tagging for example, generate the morphemes and tags in sequence, for example: 
\begin{addmargin}[1em]{2em}% 1em left, 2em right
\texttt{be>>ADP@@ha>>DET@@bayit>>NOUN ha>>DET@@lavan>>NOUN}
\end{addmargin}
\texttt{``@@"} acts as a morpheme delimiter within a word and ``>>" is the morpheme-tag delimiter. See Sec.   \ref{experiments} for more details. 
For NER and Question-Answering we can simply generate the target word forms without explicitly going through a segmentation phase.
This change in approach to using sequence-to-sequence modeling is relevant for all MRLs, and in this paper we demonstrate its effectiveness specifically for Hebrew.
% Sequence-to-sequence models, on the other hand, can simply generate the correct entity.
% \rt{the way I see things, arguments should linger from  data-->challenge-->modeling. here you kinda go the other way round and it ends up less clear}\me{I'm not sure I fully understand this. We first present the data + params probelm in hebrew nlp and only in this paragraph explain about modeling. Did I misunderstand you?}

This work thus identifies the challenge of current Hebrew NLP as a {\em three-faceted} problem: Under-parameterization, limited training data, {\em as well as} the use of a suboptimal pre-training architecture.\footnote{So far we mentioned encoder-only models as an example for  suboptimal modeling choices for MRLs, but this is also the case when using poor tokenization, small vocabularies etc.}
To address these three challenges at once, we propose using mT5 \cite{xue-etal-2021-mt5}, a large multilingual sequence-to-sequence model that was pretrained on a vast amount of multilingual data including a significant amount of Hebrew data.\footnote{It is beyond the scope of this paper to examine the factors that contributed to its improved performance, see Sec.\ \ref{limitation}.} 
To adapt classification, span prediction, and token/morpheme classification tasks to mT5's text-to-text paradigm, we propose the text-only formulations illustrated in Figure \ref{flow}.
Subsequently, we report here that this paradigm change produces empirical improvements on all tasks evaluated compared to previous state-of-the-art, some of which are dramatic, as 27.9 F1 increase in Hebrew Question-Answering.
% , ranging from small improvements on morph-syntactic tasks and up to 27.9 F1 increase on the deeper semantic task of Question-Answering. 

\section{Modeling}
We use mT5 \cite{xue-etal-2021-mt5}, a multilingual generative text-to-text version of T5 \cite{raffel2020exploring}, trained simultaneously on 101 languages.
We evaluate mT5 on all its available sizes - Small, Base, Large, XL and XXL ranging from 300M to 13B parameters.
Subsequently, we propose casting all Hebrew NLP tasks for which evaluation benchmarks exist as  text-to-text tasks: The input text is fed into the model and targets are produced in a generative manner. 

In contrast to text-to-text formulations of classification and span prediction, token classification is not as common in the literature, and specifically when the tokens consist of multiple morphemes, as is the case in MRLs.
For example, in POS tagging for MRLs, each morpheme is assigned a POS tag, unlike in non-MRLs, where tagging is a word-level task.
As a result, a generative model cannot simply generate tag predictions one after the other but it requires to first segment the text and only then label it accordingly. 
e.g., An unsatisfactory generation for ``habayit" is  \textit{DET,NOUN} as we cannot recover which morpheme belongs to which tag.
An acceptable model output, on the other hand, is \textit{ha-DET,lavan-NOUN} as we can recover that \textit{ha} was tagged with a \textit{DET} and \textit{lavan} with a \textit{NOUN}.
Throughout our experiments we tested a number of different text-to-text formulations. The best candidates are depicted in Figure \ref{flow}.
% \rt{I am not sure the modeling is clear. I think you should depict and formalize actual input and output pairs}\me{I editted this  section. Also, I'm illustrating text-2-text examples in Fig 1. Are you missing something else?}

% \begin{table}[t]
% \centering
% \resizebox{\linewidth}{!}{%
% \begin{tabular}{l|cc|cc|c|c}
% \multicolumn{1}{l|}{\multirow{3}{*}{Model}} & \multicolumn{2}{c|}{\multirow{2}{*}{ParaShoot}} & \multicolumn{2}{c|}{NEMO} & \multirow{2}{*}{BMC} & Sentiment \\
% \multicolumn{1}{l|}{} & \multicolumn{2}{c|}{} & Token & Morph. &  &  Analysis  \\
% \multicolumn{1}{l|}{} & EM & F1 & F1 & F1 & F1 & F1 \\
% \hline
% mBERT & 32 & 56.1 & 79.11 & 69.78 & 87.77 & 84.21 \\
% HeBERT & 18.2 & 36.7 & 81.13 & 77.29 & 89.41 & 87.13 \\
% AlephBERT & 26 & 49.6 & 83.62 & 77.55 & 91.12 & 89.02 \\
% ABG & - & - & 86.26 & 80.39 & - & \textbf{89.51} \\
% \hline
% mT5 - Small & 24.62 & 48.36 & 66.74 & 62.08 & 77.7 & 87.55 \\
% mT5 - Base & 36.65 & 62.97 & 74.7 & 69.12 & 85.5 & 87.25 \\
% mT5 - Large & 42.6 & 70.13 & 84.33 & 81.85 & 90.77 & 88.73 \\
% mT5 - XL & 46.15 & 73.27 & 88.65 & 84.43 & 93.01 & 88.9 \\
% mT5 - XXL & \textbf{50.37} & \textbf{77.5} & \textbf{89.86} & \textbf{88.65} & \textbf{93.29} & \textbf{89.61}
% \end{tabular}
% }
% \caption{
% mT5 outperforms previous encoder-only LMs on a variety of semantic Hebrew downstream tasks.
% }
% \label{semantic_leaderboard}
% \end{table}

% With param count
\begin{table}[t]
\centering
\resizebox{\linewidth}{!}{%
\begin{tabular}{l|c|cc|cc|c|c}
\multicolumn{1}{l|}{\multirow{3}{*}{Model}} & Param & \multicolumn{2}{c|}{\multirow{2}{*}{ParaShoot}} & \multicolumn{2}{c|}{NEMO} & \multirow{2}{*}{BMC} & Sentiment \\
\multicolumn{1}{l|}{} & Count & \multicolumn{2}{c|}{} & Token & Morph. &  &  Analysis  \\
\multicolumn{1}{l|}{} &  & EM & F1 & F1 & F1 & F1 & F1 \\
\hline
mBERT & 178M & 32 & 56.1 & 79.11 & 69.78 & 87.77 & 84.21 \\
HeBERT & 110M & 18.2 & 36.7 & 81.13 & 77.29 & 89.41 & 87.13 \\
AlephBERT & 126M & 26 & 49.6 & 83.62 & 77.55 & 91.12 & 89.02 \\
ABG & 185M & - & - & 86.26 & 80.39 & - & \textbf{89.51} \\
\hline
mT5 - Small & 300M & 24.52 & 48.71 & 66.74 & 62.08 & 77.7 & 87.55 \\
mT5 - Base & 580M & 36.65 & 62.97 & 74.7 & 69.12 & 85.5 & 87.25 \\
mT5 - Large & 1.2B & 42.6 & 70.13 & 84.33 & 81.85 & 90.77 & 88.73 \\
mT5 - XL & 3.7B & 46.15 & 73.27 & 88.65 & 84.43 & 93.01 & 88.9 \\
mT5 - XXL & 13B & \textbf{50.37} & \textbf{77.5} & \textbf{89.86} & \textbf{88.65} & \textbf{93.29} & \textbf{89.61}
\end{tabular}
}
\caption{
mT5 outperforms previous encoder-only LMs on a variety of semantic Hebrew downstream tasks.
}
\label{semantic_leaderboard}
\end{table}

\section{Experiments}
\label{experiments}

\paragraph{Goal}
The goal of this study is to assess the performance of a sequence-to-sequence large language model, specifically mT5, that was trained on a large quantity of multilingual data, compared to existing Hebrew language models, in order to showcase the efficacy of larger, well trained and thoughtfully designed models on Hebrew tasks.

\paragraph{Models}
We fine-tuned different sizes of mT5 (Small-XXL) on all Hebrew tasks in a single-task fashion for 4096 steps, with a constant learning rate of 1e-3. 
For test set evaluation, we used the best-performing checkpoint from the development set, as tasks usually converge earlier.
We compared the mT5 models against YAP \cite{more-etal-2019-joint}, mBERT \cite{devlin-etal-2019-bert}, HeBERT \cite{chriqui2021hebert}, AlephBERT \cite{seker-etal-2022-alephbert} and ABG \cite{ABG}.
We compare against YAP as it's the only available model trained on the lemmatization task. YAP's scores are produced by us. mBERT, HeBERT, AlephBERT and ABG scores are all from \citet{ABG}.
ParaShoot scores are from \citet{keren-levy-2021-parashoot}. Summary of the results can be found in Tables \ref{semantic_leaderboard}, \ref{morph_syntactic_leaderboard}.

\subsection{Tasks}

We assembled an evaluation suite of Hebrew benchmarks composed of the following tasks:
QA \cite{keren-levy-2021-parashoot}, NER \cite{bareket-tsarfaty-2021-neural, mordecai2005hebrew}, Sentiment Analysis \cite{amram-etal-2018-representations}, and the morpho-syntactic tasks of segmentation, POS tagging and lemmatization from \citet{sade-etal-2018-hebrew}, where we used the the latest dataset version, compatible with ABG.

\subsubsection{Question-Answering}

\citet{keren-levy-2021-parashoot} introduced ParaShoot, an Hebrew Question-Answering dataset which was created using the format and crowdsourcing methodology of SQuAD \cite{rajpurkar-etal-2016-squad}. We report token-level F1 and Exact Match scores as no morpheme boundaries are available.
The input is constructed by concatenating the context and question, with the output being the answer.

We conducted manual evaluation of different mT5 models on this dataset to evaluate the impact of model sizes, see details in \ref{parashoot_manualy_analysis}.

\subsubsection{Named Entity Recognition}
\citet{bareket-tsarfaty-2021-neural} created NEMO, an NER annotation for the Hebrew UD corpus \cite{sade-etal-2018-hebrew}.
As entities in Hebrew can correspond to only a subset of a word's morphemes (See ``habayit halavan" example in Sec. \ref{introduction}), the authors proposed two dataset versions: token-level, where entities correspond to white-space boundaries, similarly to BMC \cite{mordecai2005hebrew}, and morpheme-level, with morpheme-based boundaries. The authors additionally revised the common NER evaluation procedure by comparing predicted and target entities on the surface form, boundaries, and entity types, but not positions.
% \paragraph{Seq2seq formulation}
Thus, we train the sequence-to-sequence model to simply generate all of the sentence entities and tags one after the other in the input text. 

\subsubsection{Sentiment Analysis}
Correspondingly with previous work, we report F1 scores for \citet{amram-etal-2018-representations}, a sentiment analysis dataset curated by annotating Facebook user comments with positive/negative/neutral tags.\footnote{We use \citet{seker-etal-2022-alephbert} refined version which does not include leaks between split sets.}
% \paragraph{Seq2seq formulation}
Encoder input is the raw text with the decoder generates either <extra\_id\_0>, <extra\_id\_1> or <extra\_id\_2> which correspond to the positive, negative and neutral tags.
% We use special tokens to ensure that generation only requires a single token.

\subsubsection{Word Segmentation, POS Tagging and Lemmatization}
\citet{sade-etal-2018-hebrew} manually validated the UDv2 version of the Hebrew treebank resulting in a set of morpho-syntactic tasks. Aligned to previous work we report word segmentation and POS tagging.
We also evaluate our model on the lemmatization task and compare it to YAP \cite{more-etal-2019-joint}, an open-source Hebrew parser.
Correspondingly to previous work in Hebrew, we report aligned MultiSet (mset) scores.
% \paragraph{Seq2seq formulation} 
To produce the output for all these tasks we use two additional tokens: \texttt{``@@"} is the  morpheme delimiter within a word and \texttt{``>>"} is the morpheme-tag delimiter. e.g., segmenting, and POS tagging ``habayit halavan" should return \texttt{be@@ha@@bayit ha@@lavan} and \texttt{be>>ADP@@ha>>DET@@bayit>>NOUN ha>>DET@@lavan>>NOUN}, respectively.

% e.g., for the segmentation task the output is produced by each word morphemes separated with a ``@@" (e.g. be@@ha@@bayit ha@@lavan). For POS tagging and lemmatization each morpheme is assigned a tag using the ``>>" token (e.g. for POS tagging be>>ADP@@ha>>DET@@bayit>>NOUN ha>>DET@@lavan>>NOUN)

\begin{table}[t]
\resizebox{\linewidth}{!}{%
\begin{tabular}{l|c|c|c}
Model & Segmentation & POS Tagging & Lemmatization \\
\hline
YAP & 93.64 & 90.13 & 78.6 \\
mBERT & 96.07 & 93.14 & - \\
HeBERT & 97.90 & 95.80 & -\\
AlephBERT & 97.88 & 95.81 & - \\
ABG & 98.09 & 96.22 & - \\
\hline
mT5 - Small & 94.83 & 94.55 & 89.96  \\
mT5 - Base & 96.34 & 95.9 & 92.09 \\
mT5 - Large & 96.76 & 95.58 & 92.21 \\
mT5 - XL & 98.32 & 96.91 & 95.13 \\
mT5 - XXL & \textbf{98.67} & \textbf{97.46} & \textbf{95.53}
\end{tabular}
}
\caption{Morpheme-Based Aligned MultiSet (mset) Results on the UD Corpus}
\label{morph_syntactic_leaderboard}
\end{table}

\section{Results}

Our findings demonstrate a marked improvement over previously published results on existing Hebrew NLP benchmarks and by that establishing the advantage of using large multilingual sequence-to-sequence pretrained LMs such as mT5.

% QA
mT5 produces the biggest performance boost for the Question-Answering task of ParaShoot, with mT5-base already surpassing baseline models and mT5-XXL outperforming AlephBERT by 27.9 F1 points.
% The performance boost is especially drastic in this task, possibly because mT5 was pretrained in a somewhat similar span generation objective and during fine-tuning the model was not required to learn additional syntax like in the other tasks.
% \me{Can we think of something better here?}
% NER
For NER, mT5 produces better results than evaluated baselines on both datasets.  The largest performance boost comes in the morpheme-level NEMO version where mT5 learns to segment and label entities in an end-to-end fashion.

% Sentiment Analysis
For sentiment analysis, mT5 outperforms the baseline models by a small fraction, however, manual analysis we performed on mT5's best performing model show that 34\% of its errors are annotation errors and for further 30\% our annotators we not able to decide what is the correct label. We conclude that we should work towards a cleaner, more challenging sentiment analysis dataset in Hebrew.
% UD 
We report error reduction of 30.3\% and 32.8\% for the segmentation and POS tagging tasks compared to previous state-of-the-art. For the lemmatization task we report an increase of 16.93 mset F1 points compared to YAP. 
All of these are an important step towards closing the gap in morpho-syntactic tasks comparing to other languages.
% These results, together with the NER improvements possibly indicate the superiority of the sequence-to-sequence approach compared to the stacked LSTM decoder proposed in AlephBERT.
% \me{I don't like this as we can't control the model size}

\section{Related work}

HeBERT \cite{chriqui2021hebert} is the first pretrained transformer-based language model trained on Hebrew Wikipedia and OSCAR \cite{ortiz-suarez-etal-2020-monolingual} for the task of user-generated sentiment analysis.
AlephBERT \cite{seker-etal-2022-alephbert} was pretrained on the same copora in addition to a very large number of Hebrew tweets.
% This paper also proposed to stack an LSTM decoder on top of a transformer encoder, to adapt it to tasks requiring morpheme-level boundaries. 
\citet{ABG} tackled the extreme data sparseness in MRLs lexica \cite{tsarfaty-etal-2020-spmrl} by pretraining with roughly 2.5x of AlephBERT vocabulary size, leading to performance improvements. Orthogonally, \citet{keren2022breaking} proposed using char-level LMs to mitigate the same sparseness problem, albeit results were inconclusive.

\citet{xue-etal-2021-mt5} showed that mT5 outperforms baseline models on a number of multilingual datasets but did not directly evaluate on Hebrew.
Alternatively, monolingual Hebrew LM papers only compared against mBERT \cite{devlin-etal-2019-bert} as the sole multilingual baseline.

% A similar work to ours is \citet{seo-etal-2022-dog} who compared monolingual Korean LMs to mT5 and found that mT5 outperforms baselines for a variety of tasks.

\section{Conclusions}

All Hebrew LMs to date are encoder-only models, which could not directly generate morpheme sequences, and thus necessitate a specially-tuned decoder.
In this work we propose to take advantage of mT5, a publicly available multilingual large language model that was trained on a considerable amount of multilingual and Hebrew data.
Additionally the generative approach of text-to-text modeling is more aligned with the morphological challenges inherent in Hebrew and by that dispense the need for specially tuned decoders.
We fine-tuned and evaluated mT5 on a set of Hebrew downstream tasks and report that mT5 outperforms all previous baselines. Subsequently, we propose that multilingual sequence-to-sequence models should be used for Hebrew and other MRLs as an alternative for their smaller encoder-only counterparts.

\section{Limitations}
\label{limitation}
mT5, comparing to previous Hebrew LMs, is bigger, pretrained on more multiligual data, and learning to segment and tag in an end-to-end manner. While it was beyond the scope of this paper to pretrain new LMs and study which factors contributed to the improved performance, identifying these factors will be useful for determining the most effective approach for future work.
While larger mT5 models perform better than available LMs, they require more powerful hardware accelerators and take longer to train and infer. Future research should aim to achieve similar levels of performance with smaller mT5 models.
% the best performing mT5 models are bigger than available encoder-only Hebrew LMs - This may affect the GPUs required, training and inference times.
% Another limitation of our approach is that the increase of model size requires the use of larger hardware accelerators. Future work should explore ways of achieving similar levels of performance while using fewer resources. 
Additionally, the inclusion of data from 101 languages in the training of mT5 may have negatively impacted its performance on Hebrew, as some of the data may not have been relevant or beneficial to this particular language.
Future work will need to address this issue by training a monolingual Hebrew LM in order to further improve performance for Hebrew.

% The small and base mT5 models in this study are below par compared to the encoder-only baselines. , but the use of larger model sizes helped to improve performance. However, this comes at the cost of needing larger, though still widely available, GPUs/TPUs.

% mT5 tokenization is far from being optimal for Hebrew, with only a small amount of its vocabulary assigned to Hebrew wordpieces. 
% Working towards better tokenizers for Hebrew should increase the model's performance. 
% Alternatively, as suggested by \citet{keren2022breaking}, a character-base or token-free tokenization is a promising direction for non-concatenative morphology languages.

An inherent limitation in sequence-to-sequence models is that they can generate inconsistent text with respect to the input text \cite{lee2018hallucinations, rohrbach-etal-2018-object}. While potentially sensitive in different applications, a number of evaluation frameworks have been suggested to reduce the number of such ``hallucinations" \cite{honovich-etal-2021-q2, honovich-etal-2022-true}. Another limitation of our evaluation framework is that, for lack of available datasets, we did not evaluate mT5 on purely generative tasks such as summarization and paraphrasing

\section{Acknowledgements}
We thank Dan Bareket and Eylon Guetta from Bar Ilan University for their help in sharing the UD and NEMO data.

\bibliography{anthology,custom}

\begin{thebibliography}{30}
\expandafter\ifx\csname natexlab\endcsname\relax\def\natexlab#1{#1}\fi

\bibitem[{Amram et~al.(2018)Amram, Ben~David, and
  Tsarfaty}]{amram-etal-2018-representations}
Adam Amram, Anat Ben~David, and Reut Tsarfaty. 2018.
\newblock \href {https://aclanthology.org/C18-1190} {Representations and
  architectures in neural sentiment analysis for morphologically rich
  languages: A case study from {M}odern {H}ebrew}.
\newblock In \emph{Proceedings of the 27th International Conference on
  Computational Linguistics}, pages 2242--2252, Santa Fe, New Mexico, USA.
  Association for Computational Linguistics.

\bibitem[{Antoun et~al.(2020)Antoun, Baly, and Hajj}]{antoun-etal-2020-arabert}
Wissam Antoun, Fady Baly, and Hazem Hajj. 2020.
\newblock \href {https://aclanthology.org/2020.osact-1.2} {{A}ra{BERT}:
  Transformer-based model for {A}rabic language understanding}.
\newblock In \emph{Proceedings of the 4th Workshop on Open-Source Arabic
  Corpora and Processing Tools, with a Shared Task on Offensive Language
  Detection}, pages 9--15, Marseille, France. European Language Resource
  Association.

\bibitem[{Araci(2019)}]{araci2019finbert}
Dogu Araci. 2019.
\newblock Finbert: Financial sentiment analysis with pre-trained language
  models.
\newblock \emph{arXiv preprint arXiv:1908.10063}.

\bibitem[{Bareket and Tsarfaty(2021)}]{bareket-tsarfaty-2021-neural}
Dan Bareket and Reut Tsarfaty. 2021.
\newblock \href {https://doi.org/10.1162/tacl_a_00404} {Neural modeling for
  named entities and morphology ({NEMO}2)}.
\newblock \emph{Transactions of the Association for Computational Linguistics},
  9:909--928.

\bibitem[{Beltagy et~al.(2019)Beltagy, Lo, and
  Cohan}]{beltagy-etal-2019-scibert}
Iz~Beltagy, Kyle Lo, and Arman Cohan. 2019.
\newblock \href {https://doi.org/10.18653/v1/D19-1371} {{S}ci{BERT}: A
  pretrained language model for scientific text}.
\newblock In \emph{Proceedings of the 2019 Conference on Empirical Methods in
  Natural Language Processing and the 9th International Joint Conference on
  Natural Language Processing (EMNLP-IJCNLP)}, pages 3615--3620, Hong Kong,
  China. Association for Computational Linguistics.

\bibitem[{Brown et~al.(2020)Brown, Mann, Ryder, Subbiah, Kaplan, Dhariwal,
  Neelakantan, Shyam, Sastry, Askell et~al.}]{brown2020language}
Tom Brown, Benjamin Mann, Nick Ryder, Melanie Subbiah, Jared~D Kaplan, Prafulla
  Dhariwal, Arvind Neelakantan, Pranav Shyam, Girish Sastry, Amanda Askell,
  et~al. 2020.
\newblock Language models are few-shot learners.
\newblock \emph{Advances in neural information processing systems},
  33:1877--1901.

\bibitem[{Brusilovsky and Tsarfaty(2022)}]{brusilovsky2022neural}
Idan Brusilovsky and Reut Tsarfaty. 2022.
\newblock Neural token segmentation for high token-internal complexity.
\newblock \emph{arXiv preprint arXiv:2203.10845}.

\bibitem[{Chan et~al.(2020)Chan, Schweter, and
  M{\"o}ller}]{chan-etal-2020-germans}
Branden Chan, Stefan Schweter, and Timo M{\"o}ller. 2020.
\newblock \href {https://doi.org/10.18653/v1/2020.coling-main.598}
  {{G}erman{'}s next language model}.
\newblock In \emph{Proceedings of the 28th International Conference on
  Computational Linguistics}, pages 6788--6796, Barcelona, Spain (Online).
  International Committee on Computational Linguistics.

\bibitem[{Chriqui and Yahav(2022)}]{chriqui2021hebert}
Avihay Chriqui and Inbal Yahav. 2022.
\newblock Hebert \& hebemo: a hebrew bert model and a tool for polarity
  analysis and emotion recognition.
\newblock \emph{INFORMS Journal on Data Science}.

\bibitem[{Devlin et~al.(2019)Devlin, Chang, Lee, and
  Toutanova}]{devlin-etal-2019-bert}
Jacob Devlin, Ming-Wei Chang, Kenton Lee, and Kristina Toutanova. 2019.
\newblock \href {https://doi.org/10.18653/v1/N19-1423} {{BERT}: Pre-training of
  deep bidirectional transformers for language understanding}.
\newblock In \emph{Proceedings of the 2019 Conference of the North {A}merican
  Chapter of the Association for Computational Linguistics: Human Language
  Technologies, Volume 1 (Long and Short Papers)}, pages 4171--4186,
  Minneapolis, Minnesota. Association for Computational Linguistics.

\bibitem[{Guetta et~al.(2022)Guetta, Shmidman, Shmidman, Shmidman, Guedalia,
  Koppel, Bareket, Seker, and Tsarfaty}]{ABG}
Eylon Guetta, Avi Shmidman, Shaltiel Shmidman, Cheyn~Shmuel Shmidman, Joshua
  Guedalia, Moshe Koppel, Dan Bareket, Amit Seker, and Reut Tsarfaty. 2022.
\newblock \href {https://doi.org/10.48550/ARXIV.2211.15199} {Large pre-trained
  models with extra-large vocabularies: A contrastive analysis of hebrew bert
  models and a new one to outperform them all}.

\bibitem[{Hoffmann et~al.(2022)Hoffmann, Borgeaud, Mensch, Buchatskaya, Cai,
  Rutherford, Casas, Hendricks, Welbl, Clark et~al.}]{hoffmann2022training}
Jordan Hoffmann, Sebastian Borgeaud, Arthur Mensch, Elena Buchatskaya, Trevor
  Cai, Eliza Rutherford, Diego de~Las Casas, Lisa~Anne Hendricks, Johannes
  Welbl, Aidan Clark, et~al. 2022.
\newblock Training compute-optimal large language models.
\newblock \emph{arXiv preprint arXiv:2203.15556}.

\bibitem[{Honovich et~al.(2022)Honovich, Aharoni, Herzig, Taitelbaum,
  Kukliansy, Cohen, Scialom, Szpektor, Hassidim, and
  Matias}]{honovich-etal-2022-true}
Or~Honovich, Roee Aharoni, Jonathan Herzig, Hagai Taitelbaum, Doron Kukliansy,
  Vered Cohen, Thomas Scialom, Idan Szpektor, Avinatan Hassidim, and Yossi
  Matias. 2022.
\newblock \href {https://doi.org/10.18653/v1/2022.dialdoc-1.19} {{TRUE}:
  Re-evaluating factual consistency evaluation}.
\newblock In \emph{Proceedings of the Second DialDoc Workshop on
  Document-grounded Dialogue and Conversational Question Answering}, pages
  161--175, Dublin, Ireland. Association for Computational Linguistics.

\bibitem[{Honovich et~al.(2021)Honovich, Choshen, Aharoni, Neeman, Szpektor,
  and Abend}]{honovich-etal-2021-q2}
Or~Honovich, Leshem Choshen, Roee Aharoni, Ella Neeman, Idan Szpektor, and Omri
  Abend. 2021.
\newblock \href {https://doi.org/10.18653/v1/2021.emnlp-main.619} {$q^{2}$:
  {E}valuating factual consistency in knowledge-grounded dialogues via question
  generation and question answering}.
\newblock In \emph{Proceedings of the 2021 Conference on Empirical Methods in
  Natural Language Processing}, pages 7856--7870, Online and Punta Cana,
  Dominican Republic. Association for Computational Linguistics.

\bibitem[{Kaplan et~al.(2020)Kaplan, McCandlish, Henighan, Brown, Chess, Child,
  Gray, Radford, Wu, and Amodei}]{kaplan2020scaling}
Jared Kaplan, Sam McCandlish, Tom Henighan, Tom~B Brown, Benjamin Chess, Rewon
  Child, Scott Gray, Alec Radford, Jeffrey Wu, and Dario Amodei. 2020.
\newblock Scaling laws for neural language models.
\newblock \emph{arXiv preprint arXiv:2001.08361}.

\bibitem[{Keren et~al.(2022)Keren, Avinari, Tsarfaty, and
  Levy}]{keren2022breaking}
Omri Keren, Tal Avinari, Reut Tsarfaty, and Omer Levy. 2022.
\newblock Breaking character: Are subwords good enough for mrls after all?
\newblock \emph{arXiv preprint arXiv:2204.04748}.

\bibitem[{Keren and Levy(2021)}]{keren-levy-2021-parashoot}
Omri Keren and Omer Levy. 2021.
\newblock \href {https://doi.org/10.18653/v1/2021.mrqa-1.11} {{P}ara{S}hoot: A
  {H}ebrew question answering dataset}.
\newblock In \emph{Proceedings of the 3rd Workshop on Machine Reading for
  Question Answering}, pages 106--112, Punta Cana, Dominican Republic.
  Association for Computational Linguistics.

\bibitem[{Lee et~al.(2018)Lee, Firat, Agarwal, Fannjiang, and
  Sussillo}]{lee2018hallucinations}
Katherine Lee, Orhan Firat, Ashish Agarwal, Clara Fannjiang, and David
  Sussillo. 2018.
\newblock Hallucinations in neural machine translation.
\newblock \emph{NeurIPS 2018 Workshop on Interpretability and Robustness for
  Audio, Speech, and Language}.

\bibitem[{Martin et~al.(2020)Martin, Muller, Ortiz~Su{\'a}rez, Dupont, Romary,
  de~la Clergerie, Seddah, and Sagot}]{martin-etal-2020-camembert}
Louis Martin, Benjamin Muller, Pedro~Javier Ortiz~Su{\'a}rez, Yoann Dupont,
  Laurent Romary, {\'E}ric de~la Clergerie, Djam{\'e} Seddah, and Beno{\^\i}t
  Sagot. 2020.
\newblock \href {https://doi.org/10.18653/v1/2020.acl-main.645} {{C}amem{BERT}:
  a tasty {F}rench language model}.
\newblock In \emph{Proceedings of the 58th Annual Meeting of the Association
  for Computational Linguistics}, pages 7203--7219, Online. Association for
  Computational Linguistics.

\bibitem[{Mordecai and Elhadad(2005)}]{mordecai2005hebrew}
Naama~Ben Mordecai and Michael Elhadad. 2005.
\newblock Hebrew named entity recognition.
\newblock \emph{MONEY}, 81(83.93):82--49.

\bibitem[{More et~al.(2019)More, Seker, Basmova, and
  Tsarfaty}]{more-etal-2019-joint}
Amir More, Amit Seker, Victoria Basmova, and Reut Tsarfaty. 2019.
\newblock \href {https://doi.org/10.1162/tacl_a_00253} {Joint transition-based
  models for morpho-syntactic parsing: Parsing strategies for {MRL}s and a case
  study from {M}odern {H}ebrew}.
\newblock \emph{Transactions of the Association for Computational Linguistics},
  7:33--48.

\bibitem[{Ortiz~Su{\'a}rez et~al.(2020)Ortiz~Su{\'a}rez, Romary, and
  Sagot}]{ortiz-suarez-etal-2020-monolingual}
Pedro~Javier Ortiz~Su{\'a}rez, Laurent Romary, and Beno{\^\i}t Sagot. 2020.
\newblock \href {https://doi.org/10.18653/v1/2020.acl-main.156} {A monolingual
  approach to contextualized word embeddings for mid-resource languages}.
\newblock In \emph{Proceedings of the 58th Annual Meeting of the Association
  for Computational Linguistics}, pages 1703--1714, Online. Association for
  Computational Linguistics.

\bibitem[{Petroni et~al.(2019)Petroni, Rockt{\"a}schel, Riedel, Lewis, Bakhtin,
  Wu, and Miller}]{petroni-etal-2019-language}
Fabio Petroni, Tim Rockt{\"a}schel, Sebastian Riedel, Patrick Lewis, Anton
  Bakhtin, Yuxiang Wu, and Alexander Miller. 2019.
\newblock \href {https://doi.org/10.18653/v1/D19-1250} {Language models as
  knowledge bases?}
\newblock In \emph{Proceedings of the 2019 Conference on Empirical Methods in
  Natural Language Processing and the 9th International Joint Conference on
  Natural Language Processing (EMNLP-IJCNLP)}, pages 2463--2473, Hong Kong,
  China. Association for Computational Linguistics.

\bibitem[{Raffel et~al.(2020)Raffel, Shazeer, Roberts, Lee, Narang, Matena,
  Zhou, Li, Liu et~al.}]{raffel2020exploring}
Colin Raffel, Noam Shazeer, Adam Roberts, Katherine Lee, Sharan Narang, Michael
  Matena, Yanqi Zhou, Wei Li, Peter~J Liu, et~al. 2020.
\newblock Exploring the limits of transfer learning with a unified text-to-text
  transformer.
\newblock \emph{J. Mach. Learn. Res.}, 21(140):1--67.

\bibitem[{Rajpurkar et~al.(2016)Rajpurkar, Zhang, Lopyrev, and
  Liang}]{rajpurkar-etal-2016-squad}
Pranav Rajpurkar, Jian Zhang, Konstantin Lopyrev, and Percy Liang. 2016.
\newblock \href {https://doi.org/10.18653/v1/D16-1264} {{SQ}u{AD}: 100,000+
  questions for machine comprehension of text}.
\newblock In \emph{Proceedings of the 2016 Conference on Empirical Methods in
  Natural Language Processing}, pages 2383--2392, Austin, Texas. Association
  for Computational Linguistics.

\bibitem[{Rohrbach et~al.(2018)Rohrbach, Hendricks, Burns, Darrell, and
  Saenko}]{rohrbach-etal-2018-object}
Anna Rohrbach, Lisa~Anne Hendricks, Kaylee Burns, Trevor Darrell, and Kate
  Saenko. 2018.
\newblock \href {https://doi.org/10.18653/v1/D18-1437} {Object hallucination in
  image captioning}.
\newblock In \emph{Proceedings of the 2018 Conference on Empirical Methods in
  Natural Language Processing}, pages 4035--4045, Brussels, Belgium.
  Association for Computational Linguistics.

\bibitem[{Sade et~al.(2018)Sade, Seker, and Tsarfaty}]{sade-etal-2018-hebrew}
Shoval Sade, Amit Seker, and Reut Tsarfaty. 2018.
\newblock \href {https://doi.org/10.18653/v1/W18-6016} {The {H}ebrew
  {U}niversal {D}ependency treebank: Past present and future}.
\newblock In \emph{Proceedings of the Second Workshop on Universal Dependencies
  ({UDW} 2018)}, pages 133--143, Brussels, Belgium. Association for
  Computational Linguistics.

\bibitem[{Seker et~al.(2022)Seker, Bandel, Bareket, Brusilovsky, Greenfeld, and
  Tsarfaty}]{seker-etal-2022-alephbert}
Amit Seker, Elron Bandel, Dan Bareket, Idan Brusilovsky, Refael Greenfeld, and
  Reut Tsarfaty. 2022.
\newblock \href {https://doi.org/10.18653/v1/2022.acl-long.4} {{A}leph{BERT}:
  Language model pre-training and evaluation from sub-word to sentence level}.
\newblock In \emph{Proceedings of the 60th Annual Meeting of the Association
  for Computational Linguistics (Volume 1: Long Papers)}, pages 46--56, Dublin,
  Ireland. Association for Computational Linguistics.

\bibitem[{Tsarfaty et~al.(2020)Tsarfaty, Bareket, Klein, and
  Seker}]{tsarfaty-etal-2020-spmrl}
Reut Tsarfaty, Dan Bareket, Stav Klein, and Amit Seker. 2020.
\newblock \href {https://doi.org/10.18653/v1/2020.acl-main.660} {From {SPMRL}
  to {NMRL}: What did we learn (and unlearn) in a decade of parsing
  morphologically-rich languages ({MRL}s)?}
\newblock In \emph{Proceedings of the 58th Annual Meeting of the Association
  for Computational Linguistics}, pages 7396--7408, Online. Association for
  Computational Linguistics.

\bibitem[{Xue et~al.(2021)Xue, Constant, Roberts, Kale, Al-Rfou, Siddhant,
  Barua, and Raffel}]{xue-etal-2021-mt5}
Linting Xue, Noah Constant, Adam Roberts, Mihir Kale, Rami Al-Rfou, Aditya
  Siddhant, Aditya Barua, and Colin Raffel. 2021.
\newblock \href {https://doi.org/10.18653/v1/2021.naacl-main.41} {m{T}5: A
  massively multilingual pre-trained text-to-text transformer}.
\newblock In \emph{Proceedings of the 2021 Conference of the North American
  Chapter of the Association for Computational Linguistics: Human Language
  Technologies}, pages 483--498, Online. Association for Computational
  Linguistics.

\end{thebibliography}
\bibliographystyle{acl_natbib}

\appendix

\section{Running T5 on common GPUs}
\label{running_t5_on_common_gpus}
AlephBertGimmel \cite{ABG}, the largest Hebrew LM to date is roughly the same size of BERT base \cite{devlin-etal-2019-bert}, and even though mT5 is 60 times larger than AlephBertGimmel, we do not need to horizontally scale our hardware accelerators by 60 to accommodate it. As models have grown in size, hardware accelerators have also become more advanced.
T5 Small, Base and Large can all be fine-tuned in a 2016 Nvidia \texttt{P100} or 2017 Nvidia \texttt{V100} GPU accelerators. T5 XL and XXL can be fine-tuned in the 2020 Nvidia \texttt{A100} GPU, the same accelerator used for pretraining AlephBERTGimmel.

Given the widespread availability of these GPU accelerators, we argue that the mT5 models we evaluate in this work can be easily trained and deployed nowadays.

\section{Manual Evaluation of mT5 on Question-Answering}
\label{parashoot_manualy_analysis}

mT5-small performs similarly to previous state-of-the-art models on on the Question-Answering task of ParaShoot \cite{keren-levy-2021-parashoot}. We conducted manual evaluation of mT5-XXL against mT5-small, both as a way to analyse the impact of the model size, while holding other factors constant, and as a way to compare to the performance of previous state-of-the-art models.

We ran our mT5 experiments using 3 seeds with the best performing model, mT5-XXL, achieving 77.99 F1 and 50.63 EM scores. Our worst performing model, mT5-small, reached 47.67 F1 and 24.39 EM scores. From the 519 exact match prediction mT5-XXL model made, 167 of which mT5-small received F1 scored of 0. Manually evaluating the mistakes mT5-small made we conclude that for a lot of the examples the model did not manage to fully "understand the question", here mixing \textit{when} and \textit{where}:

Context:
\<l'.hr tbwst .srpt bmsgrt hm`rkh `l .srpt h.hly.t mpqd h.sb' h.srpty bmwSbh pwl lw'y lz'n.tywM lhmSyK bl.hymh l.sd .srpt h.hwpSyt>...

\noindent
Question:
\<mty h.hly.t mpqd h.sb' lhmSyK bl.hymh l.sd .srpt h.hwpSyt ? >

\noindent
The gold and mT5-XXL prediction is 
\<l'.hr tbwst .srpt bmsgrt hm`rkh `l .srpt>

The small model predicted 
\<bmwSbh pwl lw'y lz'n.tywM>.

mT5 additionally made a few hallucination errors, returning answers that were not in the original context.

Interestingly, mT5-small was able to accurately answer 49 questions that mT5-XXL was unable to, although for only 3 of these mT4-XXL received 0 F1 scored. Manually evaluating these 3 errors, 2 of them are alternative possible answers.

\end{document}